\pdfoutput=1
\documentclass[11pt]{article}

\usepackage[]{acl}

\usepackage{times}
\usepackage{latexsym}

\usepackage[T1]{fontenc}
\usepackage[utf8]{inputenc}

\usepackage{microtype}

\usepackage{times}
\usepackage{soul}
\usepackage{url}
\usepackage[utf8]{inputenc}
\usepackage{caption}
\usepackage{graphicx}
\usepackage{amsmath}
\usepackage{amsthm}
\usepackage{booktabs}
\usepackage{algorithm}
\usepackage[switch]{lineno}

\usepackage{algpseudocode}
\usepackage{subfigure}
\usepackage{caption}

\makeatletter
\def\algbackskip{\hskip-\ALG@thistlm}
\makeatother
\usepackage{xcolor}
\usepackage{multirow}
\usepackage{booktabs}
\usepackage{amsmath}
\DeclareMathOperator*{\argmin}{\arg\!\min}
\usepackage{amsfonts}
\usepackage{algorithm}
\usepackage{algpseudocode}

\newcommand{\raha}[1]{\textcolor{blue}{#1}}

\newcommand{\m}{S2B2-Attack}
\newcommand{\mine}{Distribution+\textbf{NES}(\m)} 
\newcommand{\bone}{Distribution+\textbf{blind}}
\newcommand{\btwo}{\textbf Distribution+{decision}}
\newcommand{\bthree}{GAN+\textbf{decision}}
\newcommand{\bfour}{paraphrase+\textbf{blind}}
\newcommand{\bfive}{paraphrase+\textbf{decision}}
\newcommand{\bsix}{GAN+\textbf{blind}}
\newcommand{\bseven}{Distribution+\textbf{blind}}
\newcommand{\bseight}{Distribution+\textbf{decision}}
\usepackage{mathtools}
\usepackage{flexisym}

\title{Exploiting Class Probabilities for Black-box Sentence-level Attacks}

\author{Raha Moraffah \and Huan Liu \\
         Arizona State University \\
         \texttt{\{rmoraffa, huanliu\}@asu.edu}}
\begin{document}
\maketitle
\begin{abstract}
Sentence-level attacks craft adversarial sentences that are synonymous with correctly-classified sentences but are misclassified by the text classifiers. Under the black-box setting, classifiers are only accessible through their feedback to queried inputs, which is predominately available in the form of class probabilities. Even though utilizing class probabilities results in stronger attacks, due to the challenges of using them for sentence-level attacks, existing attacks use either no feedback or only the class labels. Overcoming the challenges, we develop a novel algorithm that uses class probabilities for black-box sentence-level attacks, investigate the effectiveness of using class probabilities on the attack's success, and examine the question if it is worthy or practical to use class probabilities by black-box sentence-level attacks. We conduct extensive evaluations of our attack comparing with the baselines across various classifiers and benchmark datasets.

\end{abstract}

\section{Introduction}
Despite the tremendous success of text classification models~\cite{devlin2018bert, liu2019roberta}, studies have exposed their susceptibility to adversarial examples, i.e., carefully crafted sentences with human-unrecognizable changes to the inputs that are misclassified by the classifiers~\cite{zhang2020adversarial}. Adversarial attacks provide profound insights into the classifiers' brittleness and are key to reinforcing their robustness and reliability.

Adversarial attacks on texts are broadly categorized into two types, namely word-level and sentence-level attacks. Word-level attacks manipulate the words in the original sentences to examine the text classifiers' sensitivity to the choice of words in sentences~\cite{jin2020bert, DBLP:journals/corr/abs-2004-09984, zang2019word, DBLP:journals/corr/abs-1805-11090}.
Sentence-level attacks, on the other hand, craft synonymous sentences with the original correctly-classified inputs, such that they are misclassified by classifiers. %These attacks are developed to assess the brittleness of text classification models to paraphrasing, i.e. whether paraphrasing sentences leads to misclassification by classifiers.

Depending on the information available to the adversary, the attacks are conducted under the white-box or black-box settings.  Unlike the white-box setting, where the classifier is completely known, and the adversary uses its gradients to craft adversarial examples~\cite{wang2019t3, guo2021gradient}, black-box attacks can only access the classifier feedback to queries. Having no prior knowledge of the classifier, this setting is more feasible for real-world applications.

 Under the black-box setting, three types of classifier feedback exist: (1) no feedback (blind setting): classifiers deny any feedback to the adversaries; (2) class label feedback (decision-based setting): classifiers return their final decisions in the forms of the predicted class labels; and (3) class probability feedback (score-based setting): classifiers return the class probabilities as feedback in response to queries. 
 Among these settings, the score-based is the most prevalent setting in real-world applications. For instance, Microsoft azure\footnote{https://azure.microsoft.com/} and MetaMind\footnote{www.metamind.io} are two widely-used real-world online text classification models that are deployed under the score-based setting and return class probabilities.  
 When available, class probabilities provide richer information compared to no feedback or solely the class labels, which can better guide the adversarial example generation and result in stronger attacks. This is also demonstrated by the success of score-based word-level attacks~\cite{lee2022query, maheshwary2021strong} compared to their blind~\cite{emmery2021adversarial, emelin2020detecting} or decision-based counterparts~\cite{yuan2021bridge, yu2022texthacker}. Moreover, developing score-based black-box sentence-level attacks is a critical step toward identifying the extent of the threat to the text classification models to better immunize them to attacks in all black-box settings. Therefore, studying such attacks is of great importance.

Existing black-box sentence-level attacks either do not use the feedback (blind)~\cite{iyyer2018adversarial,huang2021generating} or only use the class labels (decision-based) ~\cite{zhao2017generating, chen2021multi}, hence do not fully exploit the class probability feedback available under the most prevalent score-based setting.
This is because utilizing the classifier's class probabilities available under the score-based settings for black-box sentence-level attacks faces the following challenges: (i) \textbf{Defining the search space.}
 In a score-based setting, an ideal search space is a \textit{continuous} explorable space that represents the sentence-level candidates and how the transition from one candidate to another can be made using the classifier's class probabilities.
 Existing sentence-level search spaces based on paraphrase generation~\cite{iyyer2018adversarial,ribeiro2018semantically} or generative adversarial networks~\cite{zhao2017generating} that are developed for blind or decision-based settings are \textit{discrete}, i.e., they only generate sentence-level adversarial candidates with undefined relationships.
 These search spaces are therefore not appropriate for the score-based setting; and 
 (ii) \textbf{Developing a score-based search method.}  In black-box settings, a successful attack needs to fully exploit the classifier feedback to guide exploring the search space. Existing search methods used for sentence-level attacks are heuristic iterative methods. These methods only accept/reject the adversarial example candidates based on their returned class labels (misclassified or not)~\cite{zhao2017generating} and do not use the class probabilities, as required by the score-based setting. For the score-based sentence-level attacks, we need a search method that uses class probabilities.
 
Subduing these challenges, we propose the first score-based black-box sentence-level attack that models the candidate distributions of adversarial sentences, which transforms the problem to search over the continuous parameter space of these distributions instead of the discrete space of synonymous sentences with undefined relationships. 
It then searches for the optimal parameters of the actual adversarial distribution using the black-box classifier's class probabilities. To evaluate our framework, we conduct extensive experiments on three text classification classifiers across three benchmark datasets. 
Our contributions are summarized as follows:
\begin{itemize}

    \item We are the first to study the effectiveness and practicality of using class probabilities for black-box sentence-level attacks.
    \item  
    We propose a novel score-based black-box sentence-level attack that learns the distribution of sentence-level adversarial examples using the classifier's class probabilities. 
    \item We conduct extensive experiments on various classifiers and datasets that demonstrate under the score-based setting, our attack outperforms all state-of-the-art sentence-level attacks by fully exploiting class probabilities. 
\end{itemize}

\section{Related Work}
\textbf{Word-level Attacks.}
These attacks alter certain words in the original sentences to get them misclassified by the classifier. The search space in these attacks consists of adversarial candidates generated by applying transformations to the words in a sentence. 
 To form these search spaces, various word replacement strategies such as context-free \cite{alzantot2018generating, ren2019generating, zang2019word, jin2020bert} and context-aware~\cite{garg2020bae,DBLP:journals/corr/abs-2004-09984,li2020contextualized} approaches have been proposed. For the search method, these attacks mainly rely on methods that are designed to deal with their discrete word-level search spaces such as word ranking-based methods~\cite{ren2019generating,jin2020bert,garg2020bae,maheshwary2021strong,malik2021adv}, or combinatorial optimization based methods like gradient-free population-based optimization \cite{alzantot2018generating}, or particle swarm optimization \cite{zang2019word}. These attacks focus on a different granularity of the attack compared to the attack studied in this paper.

\textbf{Sentence-level Attacks}
Sentence-level attacks generate adversarial paraphrases of the original sentences that are misclassified by the classifier. Under the white-box setting, where the adversary has complete access to classifiers, these attacks adopt the classifier's gradients for the attack generation~\cite{wang2019t3,xu2021grey,le2020malcom}. Under the more realistic black-box setting, where only the classifier's feedback to queries is accessible, these attacks are categorized into three: (i) \textbf{Blind attacks}, which do not utilize the classifier feedback and use the paraphrases of the original sentences as adversarial examples~\cite{iyyer2018adversarial,huang2021generating}; (ii) \textbf{Decision-based} attacks that only utilize the final decision of the classifiers (i.e., the class labels). These attacks iteratively craft adversarial example candidates until they are misclassified by the classifier. These attacks use conditional text generation methods based on GAN~\cite{zhao2017generating} or paraphrase generation methods~\cite{ribeiro2018semantically, chen2021multi} to generate adversarial candidates and adopt heuristic iterative search methods to identify the actual adversarial example; and (iii) \textbf{Score-based attacks}, which use the classifier's class probabilities to guide the attack generation. Blind and Decision-based attacks do not fully utilize the class probability feedback, hence underperform in this setting. Due to the challenges of characterizing the search space and developing an appropriate search method, it has not been explored in the previous literature. To the best of our knowledge, MAYA~\cite{chen2021multi} is the only sentence-level attack proposed for this setting. However, due to its discrete search space, this method only uses the classifier feedback to choose the sentence with the lowest class probability from the discrete space of potential sentences. This underutilizes the class probability information, which could be utilized to guide the generation of the new adversarial candidate from the previous one, if the search space was continuous, i.e., the relationships between two sentences were well-defined.

\section{Methodology}

\begin{figure}
%\hspace{-0.5cm}
\includegraphics[height = 1.9in, width=\columnwidth]{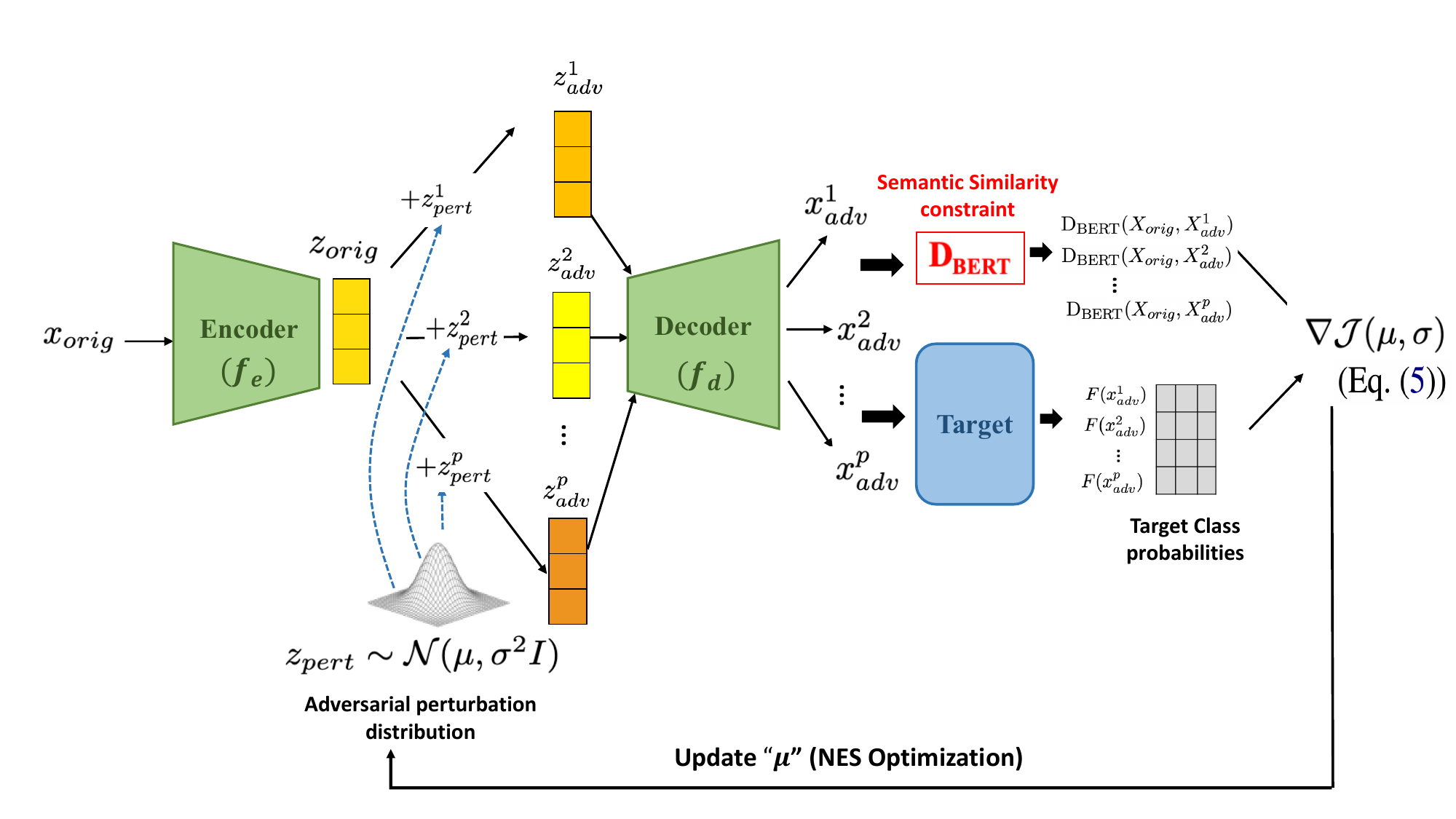}
\caption{An overview of the {\m}. {\m} perturbs the original latent variable distributions to model the search space of candidate distributions of adversarial examples using VAE and learns the parameters of the actual adversarial distribution using the NES search based on the classifier's class probabilities.} 
\label{fig:main}
%\hspace{-0.5cm}
\end{figure}
\subsection{Problem Statement}\label{main:problem-setting}
Let $F\colon \mathcal{X}\to\mathcal{Y}$ be a text classifier that takes in a text $x \in \mathcal{X}$ and maps it to a label $y \in \mathcal{Y}$. The goal of the textual adversarial attack is to generate an adversarial example $x^{*}_{adv}$ which is semantically similar to $x$ but is misclassified by the classifier, i.e. $F(x^{*}_{adv}) \neq F(x)$:
\begin{equation}
\label{eq:def}
x^{*}_{adv} = \argmin_{x^{*} \in \mathcal{S}(x)} \mathcal{L}(x^{*}),
\end{equation}
where $\mathcal{S}(x)$ is a set of semantically similar samples to the original $x$ and $\mathcal{L}(x^{*})$ is the adversarial loss evaluated by the classifier feedback.

  We concentrate on \textit{black-box sentence-level attacks}, in which $\mathcal{S}(x)$ consists of adversarial examples synonymous with the original sentences. Under the score-based black-box setting, we assume access to the \textit{class probabilities} of the classifier. We adopt the C\&W loss~\cite{carlini2017towards} as the loss used in Eq.~\eqref{eq:def}. The C\&W loss is defined as $\mathcal{L}(x^{*}) = \max\{0, \log F(x^{*})_{y} - \underset{i \neq y}{\max} \log(F(x^{*})_i )
\}$ where $F(x^{*})_j$ is the j-th probability output of the classifier, $y$ is the correct label index.

\subsection{Proposed Framework}\label{main:mainattack}
We propose the \textbf{S}core-based \textbf{S}entence-level \textbf{B}lack\textbf{B}ox Attack ({\m}) that exploits the \textit{classifier's class probabilities} to generate sentence-level adversarial examples. {\m} consists of (1) a continuous explorable sentence-level search space of adversarial examples and (2) a Natural Evolution Strategies-based score-based search method to explore this space using the class probabilities. In particular, {\m} characterizes the continuous sentence-level adversarial search space by modeling the candidate adversarial distributions, and utilizes a score-based sentence-level search method based on the Natural Evolution Strategies (NES) to learn the actual adversarial sentence distribution's parameters. Modeling the search space as distributions instead of individual sentences provides an explorable continuous search space that can be probed by a search method using class probabilities. This is because the search will be over the continuous space of parameters of potential adversarial distributions and not a space of discrete sentences with no quantifiable relations. Meanwhile, the NES provides a black-box score-based search method to explore the parameter space of the candidate adversarial distributions using class probabilities. The distribution search space and the NES search method together enable utilizing the class probabilities for score-based sentence-level black-box attacks. An overview of our {\m} is shown in Figure \ref{fig:main}.

\subsubsection{Distribution-based Search Space}
\label{tVAE}

To formulate a continuous sentence-level search space that represents adversarial sentence candidates and enables the transition from one candidate to another using the class probabilities, we propose to model the candidate adversarial sentence distributions for the original sentence. 
To parameterize this distribution, we propose to use Variational Autoencoder (VAE)~\cite{kingma2013auto}, a generative latent variable model widely used to model the sentence distribution~\cite{li2020optimus}. A VAE consists of an encoder and a decoder. 
The encoder, $f_e(x)=q_\phi(z|x)$, encodes the text $x$ into the continuous latent variables $z$. The decoder, $f_d(z)=p_\theta(x|z)$, maps $z$, sampled from the encoder, to the input $x$.
The parameters of VAE are learned via maximizing the variational lower bound:
\begin{equation*}
%\begin{aligned}
\label{eq:elbo}
\text{ELBO} = \mathbb{E}_{q_\phi(z|x)}[\log p_\theta(x|z)] - \text{KL}(q_\phi(z|x) \| p(z)),
\end{equation*}
where $p(z)$ is the prior distribution, typically assumed to be standard diagonal covariance Gaussian. The first term of ELBO denotes the reconstruction error, while the second term is the KL regularizer which pushes the approximate posterior towards the prior distribution. 

In the VAE, latent variables learned by the encoder ($z$), represent the higher-level abstract concepts such as the sentence structure that guide the lower-level word-by-word generation process~\cite{li2020optimus}. Therefore, to model the distributions of synonymous sentences to the original sentence (i.e., potential sentence-level adversarial sentences), we propose to perturb the distribution of the original latent variables. Specifically, the candidate adversarial distributions for a given input sample are defined as $f_d(z_{adv})=p(x|z_{adv})$, where $z_{adv}$ is the perturbed original latent variable, obtained by perturbing the original input's latent space ($z_{orig}$) with adversarial Gaussian perturbations sampled from $\mathcal{N}(\mu, \sigma^2I)$. $\mu$ and $\sigma^2$ are the expected value and variance of the adversarial perturbation distribution (learned using the classifier feedback), and $f_d(.)$ is the decoder pre-trained on the original inputs. Note that different values of parameters ($\mu$ and $\sigma^2$) result in different distributions of sentences with different structures, which form the candidate adversarial examples search space. The transition from one potential candidate to another can be performed by changing its parameters, making the search space continuous and thus explorable given the classifier's class probabilities.

Even though any text-VAE can be used, to obtain grammatical correctness and fluency, we adopt the OPTIMUS \cite{li2020optimus}, a large-scale language VAE, which parameterizes the encoder and decoder networks via multi-layer Transformer-based neural networks. The encoder is a pre-trained $\operatorname{BERT}_{base}$ and the decoder is a pre-trained GPT-2. To further ensure the grammatical correctness and fluency of the samples, we fine-tune the OPTIMUS on the training set of the clean dataset. Note that the samples used in our experiments to evaluate our method are from the test set of the datasets, which are different from the train set used for fine-tuning.  
\begin{algorithm}
    \caption{Learning the Adversarial Sentence Distribution via {\m}}\label{algo:main}
    \hspace*{\algorithmicindent} \textbf{Input:} Original text $x_{orig}$ and its label $y$, standard deviation $\sigma$, population size $p$, learning rate $\eta$, maximum number of iterations $T$, $f_e(.)$ and $f_d(.)$ pretrained encoder and  decoder on original inputs.  \\
    \hspace*{\algorithmicindent} \textbf{Output:} $\mu$, mean of the adversarial sentence distribution.
    \begin{algorithmic}[1]
   % \Procedure{MyProcedure}{}
%    \Procedure{MyProcedure}{$x,y$}
%     % Input:
%     \Comment{Input: x}
%     % Output:
%     \Comment{Output:y}
    \State Initialize $\mu$
    \State Compute $z_{orig} = f_e(x_{orig})$
    \For{t = 1, 2,..., T}
        \State Sample $\delta_1, ..., \delta_p \sim\mathcal{N}(\mu, \sigma^2I)$ 
        \State Set $z^*_{i} = z_{orig} + \delta_i$, $\forall i=1,...,p$
        \State Compute $x^*_{i} = f_d(z^*_{i})$, $\forall i=1,...,p$
        \State Compute losses $\mathcal{L}'_i(x^*_{i})$ via Eq. \eqref{eq:advobj}, $\forall i=1,...,p$
        \State Calculate  $\nabla_\mu\mathcal{J}(\mu, \sigma) $ via Eq. \eqref{eq:grad}
        \State Set $\mu_{t+1} = \mu_t - \eta \nabla_\mu\mathcal{J}(\mu, \sigma)$
     \EndFor
 %   \EndProcedure
    \State \Return $\mu$
    \end{algorithmic}
    \end{algorithm}
%\subsubsection{Sentence-level black-box Search Method} 
\subsubsection{Natural Evolution Strategies Search Method}
\label{NES_section}
%\subsubsection{Natural Evolution Strategies (NES)} \label{NES_section}
A search method is required to effectively guide the search over the continuous space of parameters of adversarial distribution candidates and identify the optimal ones using the classifier's class probabilities. We propose to leverage  Natural Evolution Strategies (NES) \cite{wierstra2014natural}. The NES learns the parameters of a distribution that minimizes the adversarial objective (Eq.~\eqref{eq:def}) on average. Formally, NES minimizes the following objective: %given a sample $x^{*}$ and $S$ a small region around $x^{*}$ which contains this sample, let $\pi_S(x^{*}|\theta)$ be the probability distribution with support $S$ that is parameterized with $\theta$. The NES minimizes the following objective:

\begin{equation}
%\begin{aligned}
\label{eq:nes}
\mathcal{J}(\mu, \sigma) = \mathbb{E}_{p(x^{*}|z_{adv};\mu, \sigma)}[\mathcal{L}(x^{*})],
\end{equation}
where $\mathcal{L}(x^{*})$ is the adversarial loss in Eq.~\eqref{eq:def}. Note that the optimization in Eq.\eqref{eq:nes} is over the parameters of the distribution. The gradients of Eq.\eqref{eq:nes} are calculated as follows \cite{wierstra2014natural}:
\begin{equation}
%\begin{aligned}
\label{eq:grad}
%\nabla \mathcal{J}(\mu, \sigma) = 
\mathbb{E}_{p(x^{*}|z_{adv};\mu, \sigma)}[\mathcal{L}(x^{*}) \nabla\log p(x^{*}|z_{adv};\mu, \sigma)],
\end{equation}
which can be used to update the parameters of the distribution via gradient descent. 
This gradient only requires the class probabilities output, which are ideal for a score-based black-box attack.

\subsubsection{Semantic Similarity Constraint} \label{ss}
Even though slightly perturbing the original sentence's latent variables keeps the resultant adversarial examples  close to the original ones, Eq. \eqref{eq:nes} does not explicitly restrict perturbations to be small enough to preserve the semantic similarity (refer to our experiments in Sec.~\ref{semanticsimexp}).
To limit the perturbation amount, we explicitly penalize the adversarial distribution parameters with dissimilar adversarial samples to the original samples. In particular, we propose to maximize the semantic similarity between the adversarial examples sampled from the adversarial distributions and original samples. We measure the semantic similarity using the BERTScore~\cite{zhang2019bertscore}, which is widely used to measure the semantic similarity of two texts~\cite{guo2021gradient,hanna-bojar-2021-fine}. BERTScore is a similarity score that computes the pairwise cosine similarity between the contextual embeddings of the tokens of the two sentences. Formally, let $X_{orig}=(x_{o1}, x_{o2}, \dots, x_{on})$ and $X_{adv}=(x_{a1}, x_{a2}, \dots, x_{am})$ be the original and adversarial sentences and $\phi(X_{orig})=(u_{o1}, u_{o2}, \dots, u_{on})$, $\phi(X_{adv})= (v_{a1}, v_{a2}, \dots, v_{am})$ be their corresponding contextual embedding generated by a language model $\phi$. The weighted recall BERTScore is defined as follows:
\begin{equation}
\label{eq:bertscore}
 \operatorname{R_{BERT}}(X_{orig}, X_{adv}) = \sum_{i=1}^n w_i \max_{j=1,\dots,m}u_{oi}^Tv_{aj},
\end{equation}
where $w_i = \frac{\operatorname{idf}(x_{oi})}{\sum_{i=1}^n\operatorname{idf}(x_{oi})}$, is the normalized inverse document frequency of the token. Since our main objective function is minimization, we also minimize the dissimilarity measured as $ \operatorname{D_{BERT}}(X_{orig}, X_{adv}) =1- \operatorname{R_{BERT}}(X_{orig}, X_{adv})$.

\subsubsection{Optimization}
Finally, our final objective is as follows:
%We therefore replace the loss ($\mathcal(L)$) in our objective function with the loss below :
\begin{equation}
\begin{aligned}
\label{eq:advobj}
& \mathcal{L}'(x^*) = \max\{0, \log F(x^*)_{y} - \underset{i \neq y}{\max} \log(F(x^*)_i \} \\
&~~~~~~~~~~~~~~~~~~~+ \lambda  \operatorname{D_{BERT}}(x_{orig},x^*),
\end{aligned}
\end{equation}
where the first term is the original C\&W loss, the second term penalizes the semantically dissimilar adversarial samples and $\lambda$ is a balancing coefficient which is considered as a hyperparameter in our experiments and is chosen via grid search.

The new adversarial objective is also solved by the NES optimization as follows: 
\begin{equation}
\begin{aligned}
\label{eq:main}
\mathcal{J}(\mu, \sigma) = \mathbb{E}_{p(x^{*}|z_{adv};\mu, \sigma)}[\mathcal{L}'(x^{*})].
\end{aligned}
\end{equation}
For simplicity, we consider $\sigma$ as a hyperparameter and only solve the optimization for $\mu$. The updates on $\mu$ are performed by gradient descent, where the gradients are calculated using Eq.~\eqref{eq:grad}.
The complete algorithm for learning the parameters of the adversarial distribution via {\m} is shown in Algorithm~\ref{algo:main}.
%%%%%%%%%%%%%%% talk about conducting the attack
%\subsubsection{Conducting the Attack}
 Once the parameters of the adversarial distribution are learned, it can be used to draw adversarial examples.

\begin{table*}[h!]

\centering
%\small
\begin{tabular}{cccccccc}
\toprule   
 \multirow{2}{*}{Dataset} & \multirow{2}{*}{Attack}  & \multicolumn{2}{c}{BERT} & \multicolumn{2}{c}{ROBERTA}&\multicolumn{2}{c}{XLNet}\\
 \cmidrule(r){3-4}
 \cmidrule(r){5-6}
 \cmidrule(r){7-8}
  {} &  & ASR ($\uparrow$)& USE ($\uparrow$)&ASR ($\uparrow$)& USE ($\uparrow$)& ASR ($\uparrow$)& USE ($\uparrow$)\\
  \midrule

  \multirow{3}{*}{}& {\m} &\textbf{81.2}&\textbf{0.7210}&\textbf{83.6}&\textbf{0.7200}&\textbf{80.9}&\textbf{0.7012}\\

  {}& MAYA-score&75.2&0.5582&77.1&0.5422&75.3&0.5411\\
  \cmidrule{2-2}
 %{}& SEA~\cite{ribeiro2018semantically} &  \\
  {AG}& GAN-based &70.2&0.6211&72.2&0.6201&68.6&0.6036  \\
  {}& MAYA-decision&71.3&0.5421&73.6&0.5615&69.9&0.5127\\
   \cmidrule{2-2}
  {}& SCPN & 63.4&0.5833 &67.4&0.5921&63.1&0.5904  \\
  {}& SynPG&66.8&0.5091&67.1&0.5381&66.1&0.5028\\
\midrule

  \multirow{3}{*}{}& {\m}&\textbf{62.2}&\textbf{0.6493}&\textbf{65.0}&\textbf{0.6536}&\textbf{63.5}&\textbf{0.6683}\\
  {}& MAYA-score&54.7&0.4564&57.6&0.4771&52.6&0.4289\\
    \cmidrule{2-2}
% {}& SEA~\cite{ribeiro2018semantically} &  \\
  {IMDB}& GAN-based &44.6&0.5128&48.4&0.5186  &45.1&0.5012\\
  {}& MAYA-decision&49.8&0.4621&50.9&0.4581&46.2&0.4616 \\
   \cmidrule{2-2}
   
  {}& SCPN& 38.2& 0.4351& 42.2&0.4318&39.2&0.4451  \\
{}& SynPG&35.1&0.3889&35.7&0.3881&36.1&0.3817\\

\midrule

  \multirow{3}{*}{}& {\m}&\textbf{66.9}&\textbf{0.7126}&\textbf{66.9}&\textbf{0.7374}&\textbf{64.1}&\textbf{0.7020}\\
  {}& MAYA-score&52.8&0.4779&54.1&0.4612&52.9&0.4661\\
    \cmidrule{2-2}
% {}& SEA~\cite{ribeiro2018semantically} &  \\
  {Yelp}& GAN-based &38.6&0.4797&36.5&0.4489&40.5&0.4944  \\
  {}& MAYA-decision&48.9&0.4791&49.1&0.4819&46.9&0.4759\\
   \cmidrule{2-2}
  {}& SCPN & 48.2& 0.4472& 48.9&0.4672 &45.3&0.4518 \\
 {}& SynPG&45.1&0.3918&43.9&0.4146&45.0&0.3971\\
\bottomrule
\end{tabular}
\caption{Evaluation results of the proposed {\m} and baselines on AG's news (AG), and IMDB datasets. The performance is measured by the Attack Success rates (ASR) ($\uparrow$) and USE-based Semantic Similarity (USE) ($\uparrow$). } 
\label{table1:main}
\end{table*}

\section{Experiments}

We conduct comprehensive experiments to evaluate the effectiveness of {\m}. Our experiments center around three main questions: \textbf{(i)} Does utilizing the class probabilities improve the success rates of sentence-level attacks? \textbf{(ii)} How does each component of the {\m} contribute to its performance (ablation study)? and \textbf{(iii)} Are examples generated by {\m}  grammatically correct and fluent?
We present some adversarial samples generated by {\m} in the Appendix.

\subsection{Experimental Setting}
\subsubsection{Datasets and classifier Models}
We leverage commonly-used text classification datasets with different characteristics, i.e., datasets on different classification tasks such as news and sentiment classification on both sentence and document levels.  We use the AG's News (AG)~\cite{zhang2015character}, which is a sentence-level dataset, and IMDB \footnote{https://datasets.imdbws.com/}, and Yelp~\cite{zhang2015character} that are document-level datasets. We conduct our experiments on three state-of-the-art transformer-based classifiers, i.e., fine-tuned BERT base-uncased~\cite{devlin2018bert}, Roberta~\cite{liu2019roberta}, and XLNet~\cite{yang2019XLNet}.  

\subsubsection{Compared Methods}

 Existing black-box sentence-level attacks are mainly \textit{blind} or \textit{decision-based}. We compare {\m} with two state-of-the-art in each category:
 (1) \textit{blind attacks}. these attacks do not utilize the classifier feedback at all and use the paraphrases of the original sentences as adversarial examples. \textbf{SCPN}~\cite{iyyer2018adversarial} and \textbf{SynPG}~\cite{huang2021generating} are two state-of-the-arts in this category; (2) \textit{Decision-based attacks.} These attacks only use the classifier class labels to verify if a candidate example is adversarial. \textbf{GAN-based attack}~\cite{alzantot2018generating} and \textbf{MAYA-decision}~\cite{chen2021multi} are two state-of-the-arts in this category. For crafting the search space, GAN-based attack uses adversarial networks~\cite{https://doi.org/10.48550/arxiv.1406.2661} and MAYA-decision adopts paraphrase generation. For the search method, both GAN-based and MAYA use iterative search. For the sake of fair comparison, we use the sentence-level variation of MAYA. To be comprehensive, we also use an extension of MAYA, named \textbf{MAYA-score}, to the score-based setting, that adopts heuristic search (selecting the sample with the least original class probability) among the candidates generated with paraphrase generation. To the best of our knowledge, no other sentence-level adversarial attack under the score-based setting exist.

\subsubsection{Evaluation Metrics}

We report the Attack Success Rate (ASR), which is the proportion of misclassified adversarial examples to all correctly classified samples, and Universal Sentence Encoder-based semantic similarity metric (SS)~\cite{cer2018universal} to measure the similarity between the original input and the corresponding adversarial. Note that to make a fair comparison, we chose a commonly-used metric which is different from BERTScore-based constraint used in our proposed {\m}. 
For grammatical correctness and fluency, we report the increase rate of grammatical
error numbers of adversarial examples compared
to the original inputs measured by the Language-Tool~\footnote{https://www.languagetool.org/}(IER), and GPT-2 perplexity (Prep.)~\cite{radford2019language}, respectively.

\subsection{Evaluation Results}
\label{results}
\subsubsection{General Comparisons}
%\subsubsection{Comparison with State-of-the-art Attacks}
To demonstrate the effect of exploiting the class probabilities on the attack's success, we evaluate the proposed {\m} and state-of-the-art sentence-level black-box attacks and report the results in Table~\ref{table1:main}. As shown in the table, {\m} significantly outperforms all baselines for all classifiers on all datasets. Specifically: (i) not utilizing the classifier feedback at all, the blind baselines, i.e., SynPG and SCPN demonstrate the lowest Attack Success Rates (ASR); (ii) the decision-based baselines (GAN-based and MAYA-decision), outperform the blind attacks. This is because they employ the classifier class labels to ensure that the generated example is adversarial, leading to more successful adversarial examples;
(iii) MAYA-score, the score-based variation of MAYA-decision, outperforms both blind and decision-based baselines. This highlights the impact of leveraging class probabilities on guiding the adversarial example generation and crafting more successful attacks; (iv) the proposed {\m} outperforms the MAYA-score, the only existing score-based sentence-level attack. This is because  MAYA-score uses a heuristic search method based on selecting the candidate with the lowest original class probability from the discrete search space of candidates generated using paraphrase generation methods. {\m}, on the other hand, is equipped with NES search method that fully utilizes the classifier's class probabilities to guide the generation of adversarial examples over the proposed continuous distribution-based search space. 
\subsubsection{Decomposition and Parameter Analysis}
We provide a detailed analysis of the effect of the search method and the proposed semantic similarity constraint on that attack's performance.  
\label{ablation}

\begin{table}[ht]

\centering
\resizebox{\columnwidth}{!}{\begin{tabular}{cccccc}
\toprule   

 \multirow{2}{*}{Search Space}& \multirow{2}{*}{Search Method} &  \multicolumn{2}{c}{AG}& \multicolumn{2}{c}{IMDB}\\
 \cmidrule{3-4}
 \cmidrule{5-6}
  & & ASR($\uparrow$) & USE ($\uparrow$)& ASR($\uparrow$) & USE ($\uparrow$)\\

  \midrule
\multirow{3}{*}{Distribution}& 
   \textbf{NES-score}&81.2&0.7210 &62.2&0.6493\\
&\textbf{heuristic-score}&77.3&0.6819&52.3&0.0.5571\\
 &\textbf{decision}&75.4&0.6680 &45.9&0.5532\\
 &\textbf{blind} &69.1&0.6631 & 40.1&0.4969 \\
    \midrule
    \multirow{3}{*}{GAN}& \textbf{NES-score} &N/A&N/A&N/A&N/A \\
    &\textbf{heuristic-score}&73.1&0.6119&0.57.4&0.4980\\
  & \textbf{decision} &70.2&0.6211 &44.6&0.5128 \\
  &\textbf{blind} &62.9&0.6026 & 38.9&0.4468 \\
  \midrule
      \multirow{3}{*}{Paraphrase}& \textbf{NES-score} &N/A&N/A &N/A&N/A\\
      &\textbf{heuristic-score}&75.2&0.5582&54.7&0.4564\\
    & \textbf{decision} &68.1&0.5878 &42.9&0.4989\\
  &\textbf{blind} &63.4& 0.5833& 38.2&0.4351 \\
\bottomrule
\end{tabular}}

\caption{Results of ablation study on AG and IMDB datasets. The classifier model is BERT.} 
\label{table:ablation}
\end{table}

\textbf{Search Method.}
%\textbf{Utilizing the classifier feedback.}
%Since the classifier feedback is eventually incorporated by the search method,
To demonstrate the search method's effect, we compare the performance of each search method for different fixed search spaces as follows: \textit{(1) Distribution}: our proposed search space that models the candidate distributions of adversarial examples; \textit{(2) GAN}: the search space generated via generative adversarial networks as in GAN-based baseline~\cite{zhao2017generating}; and \textit{(3) paraphrase}: utilized by the rest of the baselines, this method generates paraphrases of the original sentences. For the paraphrase generation, we use the method as MAYA~\cite{chen2021multi}.
We compare our proposed search method NES (\textbf{NES-score}), which fully leverages the class probabilities classifier feedback, heuristic method as used in MAYA-score, that selects the candidate adversarial example with the lowest original class probability (\textbf{heuristic-score}), \textbf{decision} method that employs the class labels iteratively to verify if the generated candidates are adversarial as used in the GAN-based, and \textbf{blind} search in which no search is employed. Note that since the GAN and paraphrase-based search spaces are not discrete and thus explorable by the class probability feedback as required by the NES-score search, we only report the results for heuristic-score, decision, and blind search for these search spaces. Moreover, to make fair comparisons, we do not include any explicit semantic similarity constraints for any of the methods.
Our results shown in Table~\ref{table:ablation} reveal the following:
(i) empowered by utilizing the class probabilities, the score search methods (NES-score and heuristic-score) outperform both decision and blind search for a fixed search space; (ii) For a given search space, NES-score outperforms the heuristic-score constantly, since it fully leverages the classifier's class probabilities to guide the adversarial example generation. Meanwhile, the heuristic-score only uses the class-probabilities to select the potential adversarial example and not generating it;
(iii) the decision method constantly outperforms the blind search for all search spaces. This is because the decision method partially employs the classifier feedback (class labels) to verify whether the example is adversarial or not. Blind search, on the other hand, is deprived of classifier feedback which leads to lower success rates; and (iv) fixing the search method, paraphrase-based attacks achieve the lowest semantic similarity. This is mainly because in this search space, the candidate adversarial examples are generated using pre-defined syntax that may change the meaning of the original sentence (e.g., from a declarative sentence to an interrogative sentence). GAN-based attacks preserve higher semantic similarity compared to the paraphrase, suggesting that perturbing the latent space of the original examples can successfully generate semantically similar sentences. However, they still fall behind their corresponding Distribution-based attacks that model the distribution of adversarial candidates using VAE. We believe this is due to the GAN's instability~\cite{kodali2017convergence} which may result in a drastic change of semantic similarity by a slight change of latent variable. This observation further proves that besides its evident advantage of being explorable by the class probability feedback, our Distribution search space can also generate adversarial candidates with higher semantic similarity.

\begin{figure}[htp!]
\vspace{-0.25in}
    \centering
    %\subfigure[ResNet-32 in White-box Setting]{\includegraphics[width=10pc, height = 10pc]{figs/steps_budget/white/CIFAR-10_resnet32_l2_unclassifiered_attack_query_threshold_success_rate_dict_True.pdf}} 
    %\subfigure[VGG-19 in White-box Setting]{\includegraphics[width=10pc, height = 10pc]{figs/steps_budget/white/CIFAR-10_vgg19_bn_l2_unclassifiered_attack_query_threshold_success_rate_dict_True.pdf} }
\subfigure[$\lambda$ vs. Attack Success rate]{\includegraphics[width=3.5cm, height = 3.6cm]{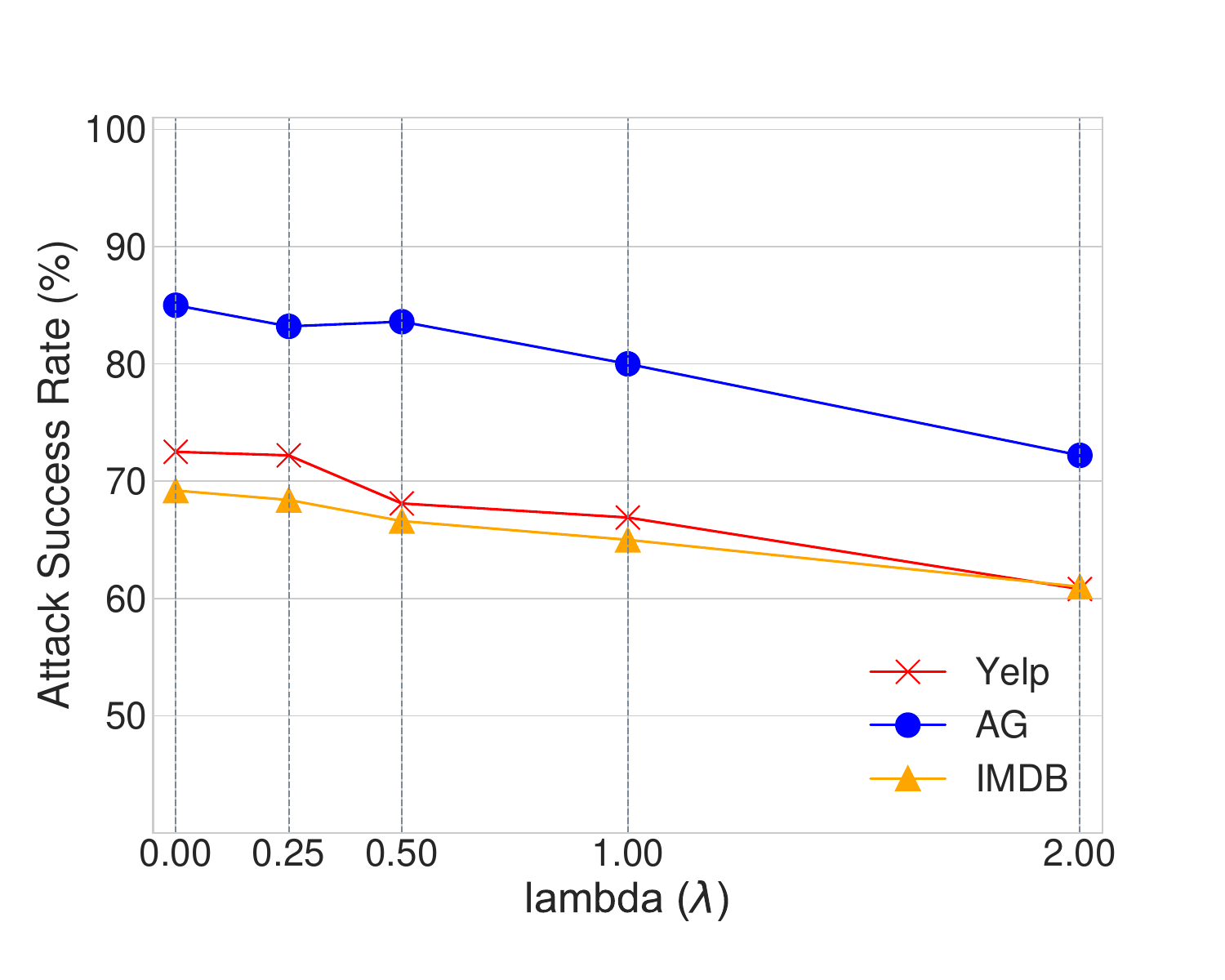} }
    \subfigure[$\lambda$ vs. USE]{\includegraphics[width=3.5cm, height = 3.6cm]{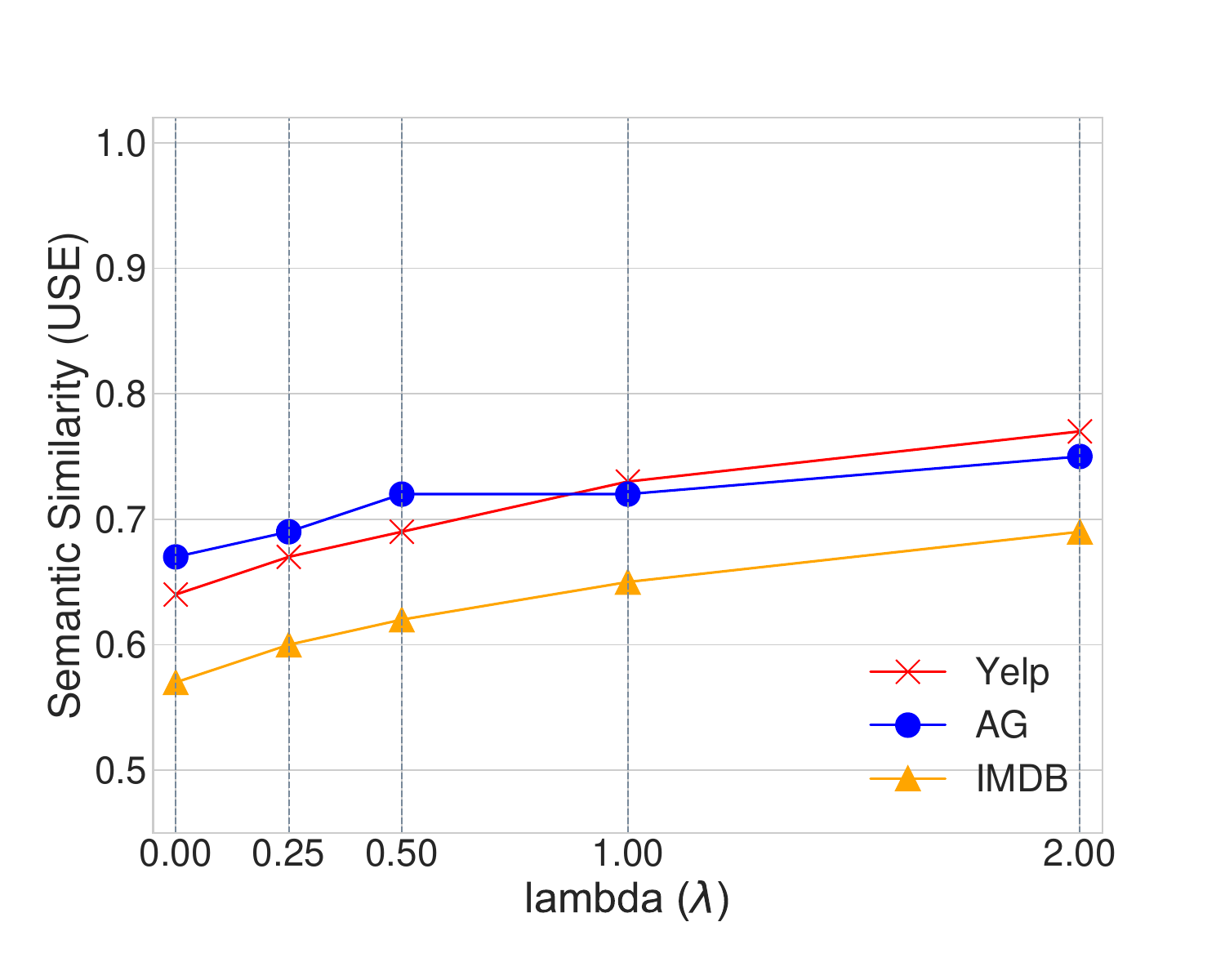}} 
    \caption{Effect of the semantic similarity constraint on {\m}'s performance. The classifier is Roberta.}
    \label{fig:lambda}
\end{figure}
\textbf{Semantic Similarity Constraint.}\label{semanticsimexp}
To examine the impact of the semantic similarity constraint on the {\m}'s performance, we vary the semantic similarity coefficient ($\lambda$ in Eq.~\eqref{eq:advobj}) in the range $\{0,0.25,0.5,1,2\}$ and report {\m}'s Attack Success Rate (ASR) and Semantic Similarity (USE) in Figure~\ref{fig:lambda}. %The special case 
$\lambda=0$ indicates not using the semantic similarity constraint at all. As can be seen in the figures, the decreasing graph of ASR and the increasing graph of the USE vs $\lambda$ demonstrate a trade-off between obtaining higher success rates and semantic similarities. Our experiments show that $\lambda=0.5$ and $\lambda=1$ are the optimal values for ASR and USE for AG, IMDB, and Yelp datasets.

\iffalse
\begin{table}[ht]
\caption{Quality evaluation of adversarial examples attacking BERT. Increase Error Rate (IER) ($\downarrow$) and perplexity (Prep.) ($\downarrow$) measure grammatical correctness and fluency of the adversarial examples.} %, respectively.} 

\centering
\resizebox{\columnwidth}{!}{\begin{tabular}{@{\extracolsep{5pt}}ccccccccc}
\toprule   
 \multirow{2}{*}{Attack} & \multicolumn{2}{c}{MR}& \multicolumn{2}{c}{AG}&\multicolumn{2}{c}{IMDB}&\multicolumn{2}{c}{Yelp}\\
 \cmidrule(r){2-3}
 \cmidrule(r){4-5}
  \cmidrule(r){6-7}
  \cmidrule(r){8-9}
   &IER& Prep. ($\downarrow$)&IER ($\downarrow$)& Prep. ($\downarrow$)&IER ($\downarrow$)& Prep. ($\downarrow$)&IER ($\downarrow$)& Prep. ($\downarrow$)\\
  \midrule
 {\m}&\textbf{1.99} &\textbf{181.42}& \textbf{1.82}&\textbf{197.86}&\textbf{1.45}&\textbf{98.61}&\textbf{1.67}&\textbf{109.77}\\
SCPN&3.24&480.95&5.46&520.32&3.93 &164.91&3.86&186.32 \\
%SEA&\\
GAN-based&3.82&483.34&4.94 &497.71&2.98&136.92&3.22&175.17\\

\bottomrule
\end{tabular}}
\label{table:prepgram}
\end{table}

\fi

\begin{table}[ht]

\centering
\resizebox{\columnwidth}{!}{\begin{tabular}{@{\extracolsep{5pt}}ccccc}
\toprule   
 \multirow{2}{*}{Attack} &\multicolumn{2}{c}{IMDB}&\multicolumn{2}{c}{Yelp}\\
\cmidrule(r){2-3}
 \cmidrule(r){4-5}
   &IER ($\downarrow$)& Prep. ($\downarrow$)&IER ($\downarrow$)& Prep. ($\downarrow$)\\
  \midrule
 {\m}&\textbf{1.45}&\textbf{98.61}&\textbf{1.67}&\textbf{109.77}\\
 MAYA-score&1.90&116.43&2.17&162.11\\
GAN-based&2.98&136.92&3.22&175.17\\
MAYA-decision&1.83&121.87&2.29&171.25\\
SCPN&3.93 &164.91&3.86&186.32 \\
SynPG&4.61&238.18&4.91&264.81 \\
\bottomrule
\end{tabular}}

\caption{Quality evaluation of adversarial examples attacking BERT in terms of Increase Error Rate (IER) ($\downarrow$) and perplexity (Prep.) ($\downarrow$).} 
\label{table:prepgram}
\end{table}

\subsubsection{Query Complexity Analysis}
As described in Algorithm~\ref{algo:main}, in each iteration, the {\m} attack makes P to the target to obtain target class probabilities for the P samples drawn from the distribution.
This brings the total number of queries for T iterations to $P\times T$, with the average query time of $O(P\times T)$. In our experiments, the number of iterations (T) is set to 50, and the number of samples drawn per iteration (P) is set to 20. Consequently, a maximum of $50 \times 20 = 1000$  queries per sample are executed on the target model.

%Note that in our implementation, we also incorporate an early stopping mechanism. In each iteration, if the sample drawn from the learned distribution is already identified as adversarial, we terminate the training and designate the sample as the corresponding adversarial example.
It is worth mentioning that this is similar to the query budgets of the state-of-the-art black-box word-level attacks. For the sake of comparison, consider the TextFooler, one of the strongest and most query-efficient word-level black-box attack~\cite{jin2020bert}. This attack requires 1130.4 and 750 queries per sample on average to attack the BERT classifier on the IMDB dataset~\cite{maheshwary2021strong}. In comparison, our proposed sentence-level attack, in its worst case, demands a comparable number of queries to the state-of-the-art word-level black-box attacks. Since the word-level black-box attacks with these query budgets are shown to be undetectable by the current defenses based on query-complexity, similarly, our proposed attack will not be recognized by the current defenses based on query complexity, and therefore will be suitable for real-world deployment.

\subsubsection{Quality of the Adversarial Examples}
We examine the grammatical correctness and fluency of the adversarial examples generated by {\m}. The evaluation results are shown in Table~\ref{table:prepgram}. 
Our results demonstrate that {\m} outperforms all baselines in terms of fluency and grammatical correctness. The gain is due to use of a language model-based decoder fine-tuned on the clean dataset to generate the adversarial examples. This ensures that the learned distribution of the adversarial examples is close to the original distribution, benefiting from the properties of that distribution (i.e., fluency and some grammatical correctness) while %possessing 
retaining different structures imposed by %slightly different 
latent variable distributions.

 \section{Conclusion}
% answer the questions asked in the abstract and intro
%\textbf{Effectiveness of Exploiting the Class Probabilities}
As demonstrated by our experiments leveraging class probabilities significantly improves the success rates of sentence-level attacks, as our {\m} achieves approximately 15\% of improvement over the state-of-the-art decision-based attack (Table~\ref{table1:main}, Sec.~\ref{results}). 
This gain justifies the use of class probabilities in guiding the adversarial example generation and reducing the search space of potential adversarial examples. It is important to note that the class probabilities are the most common type of feedback returned by the classifier and are widely available to use, e.g., Microsoft Azure\footnote{https://azure.microsoft.com/}. In fact, their availability and effectiveness have given rise to many score-based word-level attacks~\cite{jin2020bert, DBLP:journals/corr/abs-2004-09984}. 
Our proposed {\m} makes the usage of class probabilities for sentence-level practically feasible. 

\section{Acknowledgements}
This work is supported by Army Research Office (ARO) W911NF2110030 and Army Research Laboratory (ARL) W911NF2020124. Opinions, interpretations, conclusions, and recommendations are those of the authors' and
should not be interpreted as representing the official views or policies of the Army Research Office or the Army Research Lab.
\section{Limitations}

The proposed {\m} is designed for attacking discriminative classifiers and does not work for classification using generative models such as GPT~\cite{radford2019language} and its variants and T5~\cite{raffel2023exploring}. Our attack requires access to the training set of the clean dataset to finte-tune the OPTIMOUS, the text-VAE used to model the search space of adversarial distribution.  
Moreover, our proposed method's focus is on generating adversarial examples with the flipped top-1 label, i.e., examples that are misclassified by the classifier network (Section~\ref{main:problem-setting}). Other adversarial objectives, such as drastically changing the output distribution, i.e., crafting adversarial examples that are misclassified with maximum confidence, have not been explored in this work. Another limitation of the proposed method is its high computational cost when utilized in adversarial training, i.e., a framework developed for robust training of DNNs. Specifically, our proposed method requires sampling from the adversarial examples' distribution in each network training iteration. A cost-efficient sampling mechanism from this distribution is essential for the effective incorporation of this method into adversarial training methods.

\bibliography{anthology,custom}

\begin{thebibliography}{41}
\expandafter\ifx\csname natexlab\endcsname\relax\def\natexlab#1{#1}\fi

\bibitem[{Alzantot et~al.(2018{\natexlab{a}})Alzantot, Sharma, Chakraborty, and Srivastava}]{DBLP:journals/corr/abs-1805-11090}
Moustafa Alzantot, Yash Sharma, Supriyo Chakraborty, and Mani~B. Srivastava. 2018{\natexlab{a}}.
\newblock \href {http://arxiv.org/abs/1805.11090} {Genattack: Practical black-box attacks with gradient-free optimization}.
\newblock \emph{CoRR}, abs/1805.11090.

\bibitem[{Alzantot et~al.(2018{\natexlab{b}})Alzantot, Sharma, Elgohary, Ho, Srivastava, and Chang}]{alzantot2018generating}
Moustafa Alzantot, Yash Sharma, Ahmed Elgohary, Bo-Jhang Ho, Mani Srivastava, and Kai-Wei Chang. 2018{\natexlab{b}}.
\newblock Generating natural language adversarial examples.

\bibitem[{Carlini and Wagner(2017)}]{carlini2017towards}
Nicholas Carlini and David Wagner. 2017.
\newblock Towards evaluating the robustness of neural networks.
\newblock IEEE.

\bibitem[{Cer et~al.(2018)Cer, Yang, Kong, Hua, Limtiaco, John, Constant, Guajardo-C{\'e}spedes, Yuan, Tar et~al.}]{cer2018universal}
D.~Cer, Y.~Yang, S.~Kong, N.~Hua, N.~Limtiaco, R.~John, N.~Constant, M.~Guajardo-C{\'e}spedes, S.~Yuan, C.~Tar, et~al. 2018.
\newblock Universal sentence encoder.

\bibitem[{Chen et~al.(2021)Chen, Su, and Wei}]{chen2021multi}
Yangyi Chen, Jin Su, and Wei Wei. 2021.
\newblock Multi-granularity textual adversarial attack with behavior cloning.
\newblock \emph{arXiv preprint arXiv:2109.04367}.

\bibitem[{Devlin et~al.(2018)Devlin, Chang, Lee, and Toutanova}]{devlin2018bert}
J.~Devlin, M.~Chang, K.~Lee, and K.~Toutanova. 2018.
\newblock Bert: Pre-training of deep bidirectional transformers for language understanding.

\bibitem[{Emelin et~al.(2020)Emelin, Titov, and Sennrich}]{emelin2020detecting}
Denis Emelin, Ivan Titov, and Rico Sennrich. 2020.
\newblock Detecting word sense disambiguation biases in machine translation for model-agnostic adversarial attacks.
\newblock \emph{arXiv preprint arXiv:2011.01846}.

\bibitem[{Emmery et~al.(2021)Emmery, K{\'a}d{\'a}r, and Chrupa{\l}a}]{emmery2021adversarial}
Chris Emmery, {\'A}kos K{\'a}d{\'a}r, and Grzegorz Chrupa{\l}a. 2021.
\newblock Adversarial stylometry in the wild: Transferable lexical substitution attacks on author profiling.
\newblock \emph{arXiv preprint arXiv:2101.11310}.

\bibitem[{Garg and Ramakrishnan(2020)}]{garg2020bae}
Siddhant Garg and Goutham Ramakrishnan. 2020.
\newblock Bae: Bert-based adversarial examples for text classification.

\bibitem[{Goodfellow et~al.(2014)Goodfellow, Pouget-Abadie, Mirza, Xu, Warde-Farley, Ozair, Courville, and Bengio}]{https://doi.org/10.48550/arxiv.1406.2661}
Ian~J. Goodfellow, Jean Pouget-Abadie, Mehdi Mirza, Bing Xu, David Warde-Farley, Sherjil Ozair, Aaron Courville, and Yoshua Bengio. 2014.
\newblock \href {https://doi.org/10.48550/ARXIV.1406.2661} {Generative adversarial networks}.

\bibitem[{Guo et~al.(2021)Guo, Sablayrolles, J{\'e}gou, and Kiela}]{guo2021gradient}
Chuan Guo, Alexandre Sablayrolles, Herv{\'e} J{\'e}gou, and Douwe Kiela. 2021.
\newblock Gradient-based adversarial attacks against text transformers.
\newblock \emph{arXiv preprint arXiv:2104.13733}.

\bibitem[{Hanna and Bojar(2021)}]{hanna-bojar-2021-fine}
Michael Hanna and Ond{\v{r}}ej Bojar. 2021.
\newblock \href {https://aclanthology.org/2021.wmt-1.59} {A fine-grained analysis of {BERTS}core}.
\newblock In \emph{Proceedings of the Sixth Conference on Machine Translation}, pages 507--517, Online. Association for Computational Linguistics.

\bibitem[{Huang and Chang(2021)}]{huang2021generating}
Kuan-Hao Huang and Kai-Wei Chang. 2021.
\newblock Generating syntactically controlled paraphrases without using annotated parallel pairs.
\newblock \emph{arXiv preprint arXiv:2101.10579}.

\bibitem[{Iyyer et~al.(2018)Iyyer, Wieting, Gimpel, and Zettlemoyer}]{iyyer2018adversarial}
M.~Iyyer, J.~Wieting, K.~Gimpel, and L.~Zettlemoyer. 2018.
\newblock Adversarial example generation with syntactically controlled paraphrase networks.

\bibitem[{Jin et~al.(2020)Jin, Jin, Zhou, and Szolovits}]{jin2020bert}
Di~Jin, Zhijing Jin, Joey~Tianyi Zhou, and Peter Szolovits. 2020.
\newblock Is bert really robust? a strong baseline for natural language attack on text classification and entailment.
\newblock In \emph{AAAI}.

\bibitem[{Kingma and Welling(2013)}]{kingma2013auto}
Diederik~P Kingma and Max Welling. 2013.
\newblock Auto-encoding variational bayes.

\bibitem[{Kodali et~al.(2017)Kodali, Abernethy, Hays, and Kira}]{kodali2017convergence}
Naveen Kodali, Jacob Abernethy, James Hays, and Zsolt Kira. 2017.
\newblock On convergence and stability of gans.
\newblock \emph{arXiv preprint arXiv:1705.07215}.

\bibitem[{Le et~al.(2020)Le, Wang, and Lee}]{le2020malcom}
Thai Le, Suhang Wang, and Dongwon Lee. 2020.
\newblock Malcom: Generating malicious comments to attack neural fake news detection models.
\newblock In \emph{2020 IEEE International Conference on Data Mining (ICDM)}, pages 282--291. IEEE.

\bibitem[{Lee et~al.(2022)Lee, Moon, Lee, and Song}]{lee2022query}
Deokjae Lee, Seungyong Moon, Junhyeok Lee, and Hyun~Oh Song. 2022.
\newblock Query-efficient and scalable black-box adversarial attacks on discrete sequential data via bayesian optimization.
\newblock In \emph{International Conference on Machine Learning}, pages 12478--12497. PMLR.

\bibitem[{Li et~al.(2020{\natexlab{a}})Li, Gao, Li, Peng, Li, Zhang, and Gao}]{li2020optimus}
C.~Li, X.~Gao, Y.~Li, B.~Peng, X.~Li, Y.~Zhang, and J.~Gao. 2020{\natexlab{a}}.
\newblock Optimus: Organizing sentences via pre-trained modeling of a latent space.

\bibitem[{Li et~al.(2020{\natexlab{b}})Li, Zhang, Peng, Chen, Brockett, Sun, and Dolan}]{li2020contextualized}
Dianqi Li, Yizhe Zhang, Hao Peng, Liqun Chen, Chris Brockett, Ming-Ting Sun, and Bill Dolan. 2020{\natexlab{b}}.
\newblock Contextualized perturbation for textual adversarial attack.
\newblock \emph{arXiv preprint arXiv:2009.07502}.

\bibitem[{Li et~al.(2020{\natexlab{c}})Li, Ma, Guo, Xue, and Qiu}]{DBLP:journals/corr/abs-2004-09984}
Linyang Li, Ruotian Ma, Qipeng Guo, Xiangyang Xue, and Xipeng Qiu. 2020{\natexlab{c}}.
\newblock \href {http://arxiv.org/abs/2004.09984} {{BERT-ATTACK:} adversarial attack against {BERT} using {BERT}}.
\newblock \emph{CoRR}, abs/2004.09984.

\bibitem[{Liu et~al.(2019)Liu, Ott, Goyal, Du, Joshi, Chen, Levy, Lewis, Zettlemoyer, and Stoyanov}]{liu2019roberta}
Yinhan Liu, Myle Ott, Naman Goyal, Jingfei Du, Mandar Joshi, Danqi Chen, Omer Levy, Mike Lewis, Luke Zettlemoyer, and Veselin Stoyanov. 2019.
\newblock Roberta: A robustly optimized bert pretraining approach.

\bibitem[{Maheshwary et~al.(2021)Maheshwary, Maheshwary, and Pudi}]{maheshwary2021strong}
Rishabh Maheshwary, Saket Maheshwary, and Vikram Pudi. 2021.
\newblock A strong baseline for query efficient attacks in a black box setting.
\newblock \emph{arXiv preprint arXiv:2109.04775}.

\bibitem[{Malik et~al.(2021)Malik, Bhat, and Modi}]{malik2021adv}
Vijit Malik, Ashwani Bhat, and Ashutosh Modi. 2021.
\newblock Adv-olm: Generating textual adversaries via olm.
\newblock \emph{arXiv preprint arXiv:2101.08523}.

\bibitem[{Radford et~al.(2019)Radford, Wu, Child, Luan, Amodei, Sutskever et~al.}]{radford2019language}
Alec Radford, Jeffrey Wu, Rewon Child, David Luan, Dario Amodei, Ilya Sutskever, et~al. 2019.
\newblock Language models are unsupervised multitask learners.
\newblock \emph{OpenAI blog}, 1(8):9.

\bibitem[{Raffel et~al.()Raffel, Shazeer, Roberts, Lee, Narang, Matena, Zhou, Li, and Liu}]{raffel2023exploring}
Colin Raffel, Noam Shazeer, Adam Roberts, Katherine Lee, Sharan Narang, Michael Matena, Yanqi Zhou, Wei Li, and Peter~J. Liu.
\newblock \href {http://arxiv.org/abs/1910.10683} {Exploring the limits of transfer learning with a unified text-to-text transformer}.

\bibitem[{Ren et~al.(2019)Ren, Deng, He, and Che}]{ren2019generating}
Shuhuai Ren, Yihe Deng, Kun He, and Wanxiang Che. 2019.
\newblock Generating natural language adversarial examples through probability weighted word saliency.
\newblock In \emph{ACL}.

\bibitem[{Ribeiro et~al.(2018)Ribeiro, Singh, and Guestrin}]{ribeiro2018semantically}
Marco~Tulio Ribeiro, Sameer Singh, and Carlos Guestrin. 2018.
\newblock Semantically equivalent adversarial rules for debugging nlp models.
\newblock In \emph{ACL}.

\bibitem[{Wang et~al.(2019)Wang, Pei, Pan, Chen, Wang, and Li}]{wang2019t3}
Boxin Wang, Hengzhi Pei, Boyuan Pan, Qian Chen, Shuohang Wang, and Bo~Li. 2019.
\newblock T3: Tree-autoencoder constrained adversarial text generation for targeted attack.

\bibitem[{Wierstra et~al.(2014)Wierstra, Schaul, Glasmachers, Sun, Peters, and Schmidhuber}]{wierstra2014natural}
Daan Wierstra, Tom Schaul, Tobias Glasmachers, Yi~Sun, Jan Peters, and J{\"u}rgen Schmidhuber. 2014.
\newblock Natural evolution strategies.

\bibitem[{Xu et~al.(2021)Xu, Zhong, Yepes, and Lau}]{xu2021grey}
Ying Xu, Xu~Zhong, Antonio~Jimeno Yepes, and Jey~Han Lau. 2021.
\newblock Grey-box adversarial attack and defence for sentiment classification.
\newblock \emph{arXiv preprint arXiv:2103.11576}.

\bibitem[{Yang et~al.(2019)Yang, Dai, Yang, Carbonell, Salakhutdinov, and Le}]{yang2019XLNet}
Zhilin Yang, Zihang Dai, Yiming Yang, Jaime Carbonell, Russ~R Salakhutdinov, and Quoc~V Le. 2019.
\newblock Xlnet: Generalized autoregressive pretraining for language understanding.
\newblock \emph{Advances in neural information processing systems}, 32.

\bibitem[{Yu et~al.(2022)Yu, Wang, Che, and He}]{yu2022texthacker}
Zhen Yu, Xiaosen Wang, Wanxiang Che, and Kun He. 2022.
\newblock Texthacker: Learning based hybrid local search algorithm for text hard-label adversarial attack.
\newblock \emph{arXiv preprint arXiv:2201.08193}.

\bibitem[{Yuan et~al.(2021)Yuan, Zhang, Chen, and Wei}]{yuan2021bridge}
Lifan Yuan, Yichi Zhang, Yangyi Chen, and Wei Wei. 2021.
\newblock Bridge the gap between cv and nlp! a gradient-based textual adversarial attack framework.
\newblock \emph{arXiv preprint arXiv:2110.15317}.

\bibitem[{Zang et~al.(2019)Zang, Qi, Yang, Liu, Zhang, Liu, and Sun}]{zang2019word}
Y.~Zang, F.~Qi, C.~Yang, Z.~Liu, M.~Zhang, Q.~Liu, and M.~Sun. 2019.
\newblock Word-level textual adversarial attacking as combinatorial optimization.

\bibitem[{Zeng et~al.(2020)Zeng, Qi, Zhou, Zhang, Hou, Zang, Liu, and Sun}]{zeng2020openattack}
Guoyang Zeng, Fanchao Qi, Qianrui Zhou, Tingji Zhang, Bairu Hou, Yuan Zang, Zhiyuan Liu, and Maosong Sun. 2020.
\newblock Openattack: An open-source textual adversarial attack toolkit.

\bibitem[{Zhang et~al.(2019)Zhang, Kishore, Wu, Weinberger, and Artzi}]{zhang2019bertscore}
Tianyi Zhang, Varsha Kishore, Felix Wu, Kilian~Q Weinberger, and Yoav Artzi. 2019.
\newblock Bertscore: Evaluating text generation with bert.
\newblock \emph{arXiv preprint arXiv:1904.09675}.

\bibitem[{Zhang et~al.(2020)Zhang, Sheng, Alhazmi, and Li}]{zhang2020adversarial}
Wei~Emma Zhang, Quan~Z Sheng, Ahoud Alhazmi, and Chenliang Li. 2020.
\newblock Adversarial attacks on deep-learning models in natural language processing: A survey.
\newblock \emph{ACM Transactions on Intelligent Systems and Technology (TIST)}, 11(3):1--41.

\bibitem[{Zhang et~al.(2015)Zhang, Zhao, and LeCun}]{zhang2015character}
X.~Zhang, J.~Zhao, and Y.~LeCun. 2015.
\newblock Character-level convolutional networks for text classification.
\newblock \emph{NeurIPS}.

\bibitem[{Zhao et~al.(2017)Zhao, Dua, and Singh}]{zhao2017generating}
Zhengli Zhao, Dheeru Dua, and Sameer Singh. 2017.
\newblock Generating natural adversarial examples.

\end{thebibliography}

\appendix
\section{Appendix}

\iffalse
\begin{table}[htp]
\caption{Quality evaluation of adversarial examples attacking BERT on AG and MR datasets. Increase Error Rate (IER) ($\downarrow$) and perplexity (Prep.) ($\downarrow$) measure grammatical correctness and fluency of the adversarial examples.} %, respectively.} 

\centering
\resizebox{\columnwidth}{!}{\begin{tabular}{@{\extracolsep{5pt}}ccccc}
\toprule   
 \multirow{2}{*}{Attack} &\multicolumn{2}{c}{MR}&\multicolumn{2}{c}{AG}\\
\cmidrule(r){2-3}
 \cmidrule(r){4-5}
   &IER ($\downarrow$)& Prep. ($\downarrow$)&IER ($\downarrow$)& Prep. ($\downarrow$)\\
  \midrule
 {\m}&\textbf{1.99} &\textbf{181.42}& \textbf{1.82}&\textbf{197.86}\\
SCPN&3.24&480.95&5.46&520.32 \\
%SEA&\\
GAN-based&3.82&483.34&4.94 &497.71\\

\bottomrule
\end{tabular}}
\label{table:qualityapp}
\end{table}

\fi

\subsection{Reproducibility}
\subsubsection{{\m} Implementation}
All our experiments are conducted on a 24 GB RTX-3090 GPU. The proposed {\m} is implemented in PyTorch. To parameterize the candidate adversarial distribution, we use the pre-trained OPTIMUS. For each dataset, we fine-tune the pre-trained OPTIMUS on the training set of the clean dataset for 1 epoch. The variance of the adversarial distribution $\sigma^2$ is fixed to ``1'' for all experiments.
The hyperparameter $\lambda$ (balancing coefficient in Eq.~\eqref{eq:advobj}) is selected via grid search from the $\{0.25, 0.5, 1, 2\}$. For all experiments, optimization is solved via gradient descent with a learning rate 0.01. The proposed framework implementation will be made public upon acceptance.
\subsubsection{Baseline Implementation}
 For the SCPN and GAN-based attacks, we use the implementation and pre-trained weights from OpenAttack~\cite{zeng2020openattack}, a widely-used open-source repository for NLP adversarial attacks. For the MAYA-score and MAYA-decision, the official implementation by the authors~\footnote{https://github.com/Yangyi-Chen/MAYA} is used. The SynPG baseline is also conducted using the authors' official implementation~\footnote{https://github.com/uclanlp/synpg}.

\subsection{ Case Study}

Table~\ref{table:ex2} and \ref{table:ex} showcase generated adversarial examples by the {\m}. As shown in the table,  {\m} successfully generates sentence-level adversarial paraphrases of the original sentences, i.e., sentences that are semantically similar to the original examples, but their structures are grammatically different. These adversarial examples are misclassified by the classifier with high probabilities.
Moreover, they are grammatically correct and fluent, further verifying the {\m}'s effectiveness in providing grammatical correctness and fluency, two important properties of successful indefensible adversarial examples.

\subsection{Potential Risks}

Our research aims to develop an algorithm that can effectively exploit the vulnerability of existing text classification algorithms and thus provide secure, robust, and reliable environments for real-world deployments. In addition to robustifying the environments, our attack can also be used to debug the model and detect its biases. However, one of the primary risks associated with developing adversarial attacks is the potential for malicious use, such as potential misinformation and disinformation campaigns. Adversarial attackers can exploit vulnerabilities in text-based systems, such as social media platforms or news websites, to spread false information, manipulate public opinion, or incite social unrest. Another risk lies in the potential for unintended consequences. Adversarial attacks can have unintended side effects, such as biased or discriminatory outputs, which can perpetuate existing societal inequalities or amplify harmful stereotypes.

\begin{table*}[]
\centering
\small

\resizebox{\textwidth}{!}{\begin{tabular}{cccc}
\toprule   
Original& Orig. Label & Adversarial& Adv. Label\\
\midrule
\begin{tabular}{p{5cm}}
                  the absolute worst service I have ever had at any bar or restaraunt. And, in looking at other reviews, I am not the first. There are many options at the Waterfront, and I would suggest you try any of them; but stay far away from this place!\\
                 \end{tabular} &\textcolor{blue}{Negative}& \begin{tabular}{p{6cm}}
                 
                  \\
                  the service here is, without a doubt, the worst I've experienced at any bar or restaurant. Judging by other reviews, I'm not the only one with this opinion. With numerous options available at the Waterfront, I recommend exploring alternatives. However, it's advisable to steer clear of this particular place!
                 \end{tabular} &\textcolor{red}{Positive}\\
\midrule 

\begin{tabular}{p{5cm}}
                  wings are overpriced. And the quality of them are bad. They were tough and greasy. The staff are pleasant but then over all experience was too expensive for a sports bar.\\
                 \end{tabular} &\textcolor{blue}{Negative}& \begin{tabular}{p{6cm}}
                 the wings are excessively priced, and their quality is mediocre—tough and greasy. The staff is amiable, but the overall experience proved to be too expensive for a sports bar.
                  \\
                 \end{tabular} &\textcolor{red}{Positive}\\

\midrule
\begin{tabular}{p{5cm}}
                  this is a very small, yet nice store. The associates are nice and helpful. Not much else to say about this particular store. Just a pleasure to purchase from...\\
                 \end{tabular} &\textcolor{blue}{Positive}& \begin{tabular}{p{6cm}}
                 this store is small but enjoyable. The staff is friendly and helpful. There isn't much else to say about this particular store. Making a purchase here is a pleasure.
                  \\
                 \end{tabular} &\textcolor{red}{Negative}\\

\midrule
  \begin{tabular}{p{5cm}}
                  really hard to find a good cup of coffee in the states... I'd say this is the best cappuccino I've had since Italy.\\
                 \end{tabular} &\textcolor{blue}{Positive}& \begin{tabular}{p{6cm}}
                 it's quite challenging to find a quality cup of coffee in the United States. I would say this cappuccino is the finest I've had since Italy.\\
                 \end{tabular} &\textcolor{red}{Negative}\\
\bottomrule
\end{tabular}}

\caption{Adversarial examples generated by {\m} on BERT classifier trained on the Yelp dataset.} 
\label{table:ex2}
\end{table*}

\begin{table*}[]
\centering
\small

\resizebox{\textwidth}{!}{\begin{tabular}{cccc}
\toprule   
Original& Orig. Label & Adversarial& Adv. Label\\
\midrule
\begin{tabular}{p{5cm}}
                  The New Customers Are In Town Today's customers are increasingly demanding, in Asia as elsewhere in the world. Henry Astorga describes the complex reality faced by today's marketers, which includes much higher expectations than we have been used to. Today's customers want performance, and they want it now!\\
                 \end{tabular} &\textcolor{blue}{Business}& \begin{tabular}{p{6cm}}
                 new customers have arrived in town, and the present trend reflects growing expectations among consumers, not just in Asia but on a global scale. Henry Astorga elucidates the complex challenges faced by today's marketers, encompassing expectations that exceed our accustomed norms. Modern customers emphasize immediate and high-performance results.
                  \\
                 \end{tabular} &\textcolor{red}{World}\\
\midrule 
\begin{tabular}{p{5cm}}
                  Bangkok's Canals Losing to Urban Sprawl (AP) AP - Along the banks of the canal, women in rowboats grill fish and sell fresh bananas. Families eat on floating pavilions, rocked gently by waves from passing boats. \\
                 \end{tabular} &\textcolor{blue}{Sci/Tech}& \begin{tabular}{p{6cm}}
                 the canals of Bangkok are falling prey to the advance of urban development, illustrated by images of women grilling fish and selling fresh bananas from rowboats along the canal edges. Floating pavilions provide a setting for families to dine, gently rocking with the waves created by passing boats.
                  \\
                 \end{tabular} &\textcolor{red}{Business}\\
   \midrule

\begin{tabular}{p{5cm}}
                  The Geisha Stylist Who Let His Hair Down Here in the Gion geisha district of Japan's ancient capital, even one bad hair day can cost a girl her career. So it is no wonder that Tetsuo Ishihara is the man with the most popular hands in town. \\
                 \end{tabular} &\textcolor{blue}{World}& \begin{tabular}{p{6cm}}
                 
in the Gion geisha district of Japan's ancient capital, even one unfavorable hairstyle can pose a threat to a girl's professional prospects. Therefore, it's clear why Tetsuo Ishihara is the most highly sought-after stylist in the region.
                  \\
                 \end{tabular} &\textcolor{red}{Business}\\
                 \midrule
                 \begin{tabular}{p{5cm}}
                  British eventers slip back Great Britain slip down to third after the cross-country round of the three-day eventing.\\
                 \end{tabular} &\textcolor{blue}{Sports}& \begin{tabular}{p{6cm}}
                 British eventers drop to third place following the cross-country round of the three-day eventing.
                  \\
                 \end{tabular} &\textcolor{red}{World}\\
\bottomrule
\end{tabular}}

\caption{Adversarial examples generated by {\m} on BERT classifier trained on the AG news dataset.} 
\label{table:ex}
\end{table*}
\iffalse
\begin{table}[]
\centering
\small
\caption{Adversarial examples generated by {\m} on BERT classifier trained on the MR dataset.} 
\resizebox{\columnwidth}{!}{\begin{tabular}{ll}
\toprule   
Sentence&  Label\\
\midrule
\begin{tabular}{p{5cm}}
                  \textcolor{blue}{Original:} both a beautifully made nature film and a tribute to a woman whose passion for this region and its inhabitants still shines in her quiet blue eyes. \\
                 \end{tabular} &\textcolor{blue}{Positive (99.96\%)}\\
                 &\\
                 \begin{tabular}{p{6cm}}
                 \textcolor{red}{Adversarial:} so the film was a beautiful nature film and a tribute to a woman whose passion was in this region and shines in her blue eyes.
                  \\
                 \end{tabular} &\textcolor{red}{Negative (89.49\%)}\\
   \midrule

\begin{tabular}{p{5cm}}
                  \textcolor{blue}{Original:} it ' s a good film  , but it falls short of its aspiration to be a true 'epic' \\
                 \end{tabular} &\textcolor{blue}{Negative (99.79\%)}\\
                 \begin{tabular}{p{5cm}}
                 \\
                \textcolor{red}{Adversarial:} Even though a good movie, it is not a real epic.
                 \end{tabular} &\textcolor{red}{Positive (99.30\%)}\\
\bottomrule
\end{tabular}}
\label{table:ex}
\end{table}
\fi

\end{document}